\providecommand{\paperoptions}{ustc,normalcite}
\documentclass[\paperoptions]{meta}
\microtypesetup{expansion=false}

\usepackage{xspace}
\usepackage{amsmath,amssymb}
\usepackage{tabularx,booktabs,array,colortbl,longtable}
\usepackage{float}
\usepackage{wrapfig}
\usepackage{caption}
\usepackage{xcolor}
\usepackage{fvextra}
\usepackage{enumitem}
\usepackage{etoolbox}
\newcommand{\palate}{\textsc{Palate}}
\newcommand{\best}[1]{\textbf{#1}}
\newcommand{\code}[1]{\texttt{#1}}
\newcolumntype{L}[1]{>{\raggedright\arraybackslash}p{#1}}
\newcolumntype{R}[1]{>{\raggedleft\arraybackslash}p{#1}}
\setlist{nosep,leftmargin=1.5em}
\titleformat*{\paragraph}{\bfseries}

\definecolor{PromptTitle}{HTML}{E8EEF7}
\definecolor{PromptFrame}{HTML}{66758A}
\definecolor{PromptText}{HTML}{1F2937}
\newcommand{\prompttitle}[1]{%
  \par\medskip\noindent
  \colorbox{PromptTitle}{\parbox{\dimexpr\linewidth-2\fboxsep\relax}{%
    \textbf{#1}}}\par\nobreak\vspace{-1pt}}
\DefineVerbatimEnvironment{Prompt}{Verbatim}{
  breaklines=true,
  breakanywhere=true,
  frame=single,
  rulecolor=\color{PromptFrame},
  framesep=3mm,
  fontsize=\small,
  formatcom=\color{PromptText},
  tabsize=2
}

\title{Beyond Borrowed Histories: Person-Aligned User Simulation for
Interactive Role-Playing Evaluation}
\author[1,*,\mathsection]{\href{https://openreview.net/profile?id=~Yuhang_Zhu6}{Yuhang Zhu}}
\author[1,*]{\href{https://openreview.net/profile?id=~Mingxuan_Du3}{Mingxuan Du}}
\author[1,2,\ddagger]{\href{https://openreview.net/profile?id=~Benfeng_Xu1}{Benfeng Xu}}
\author[2]{\href{https://openreview.net/profile?id=~Jie_Gao8}{Jie Gao}}
\author[1,\dagger]{\href{https://openreview.net/profile?id=~Lingyun_Yu1}{Lingyun Yu}}
\author[1]{\href{https://openreview.net/profile?id=~Hongtao_Xie2}{Hongtao Xie}}
\affiliation[1]{University of Science and Technology of China}
\affiliation[2]{MetaStone Technology, Beijing, China}
\contribution[*]{Equal contribution}
\contribution[\mathsection]{Work done during the internship at MetaStone}
\contribution[\dagger]{Corresponding author}
\contribution[\ddagger]{Project lead}
\abstract{Role-playing agents (RPAs) have become one of the most important consumer
applications of large language models. Users engage in multi-turn
conversations with RPAs for experiences such as emotional comfort, making
reliable evaluation essential for measuring capability, comparing systems,
and guiding further improvement. Existing benchmarks, however, typically
require an RPA to continue a fixed dialogue history and then evaluate the
continuation using a fixed rubric detached from the user. We identify and
empirically demonstrate two limitations of this design. First, an RPA's output
is shaped by the preceding dialogue history, preventing a scientifically
grounded assessment of its role-playing ability in real multi-turn settings.
Second, user experience varies substantially across individuals, and
conventional fixed rubrics need not align with user satisfaction. We therefore
introduce \textbf{PALATE} (\textbf{P}erson-\textbf{A}ligned
\textbf{L}LM-Simulated-User \textbf{A}ssessment with
\textbf{T}ailored \textbf{E}valuation), a scalable RPA benchmark built on user
simulators. PALATE is
accompanied by a pool of 300 character profiles. Its main evaluation trains
five per-user simulators and lets them engage candidate RPAs in free-form,
multi-turn conversations over a pre-frozen panel of character profiles.
Alongside a general quality rubric, we construct personalized rubrics to
measure user satisfaction; on held-out annotated data, the personalized
rubrics show higher agreement with human judgments than the general rubric.
In the main evaluation of 16 candidates, PALATE separately characterizes
generic turn quality, long-horizon session capability, and per-user experience
on multi-turn trajectories co-constructed by each candidate. It thereby
produces interpretable evaluations of specific user--RPA pairs rather than
compressing systems into a single user-independent ranking.
}
\metadata[Resources]{\href{https://github.com/Zhuyh1139/PALATE}{PALATE repository}.}
\hypersetup{
  pdfauthor={Yuhang Zhu, Mingxuan Du, Benfeng Xu, Jie Gao, Lingyun Yu, Hongtao Xie},
  pdftitle={Beyond Borrowed Histories: Person-Aligned User Simulation for Interactive Role-Playing Evaluation}
}

\begin{document}
\maketitle

\section{Introduction}

Role-playing agents (RPAs) provide interactive storytelling, companionship,
and emotionally engaging conversation. Their value often unfolds over
multiple turns, and deployed systems already serve millions of users
\citep{chai2023}. As these systems enter large-scale real-world use, reliable
evaluation becomes essential for measuring capability, comparing systems,
and guiding further improvement.

Unlike general-purpose task assistants, the object being evaluated is not an
isolated response but a dialogue jointly constructed by the RPA and its user.
Such dialogues rarely have a uniquely correct answer, and conventional
automatic metrics correlate poorly with human judgments in open-ended dialogue
\citep{liu2016hownot}. Quality must instead be established through measures of
character fidelity, narrative development, and user experience.

\begin{wrapfigure}{r}{0.49\textwidth}
\centering
\vspace{-0.8\baselineskip}
\includegraphics[width=\linewidth]{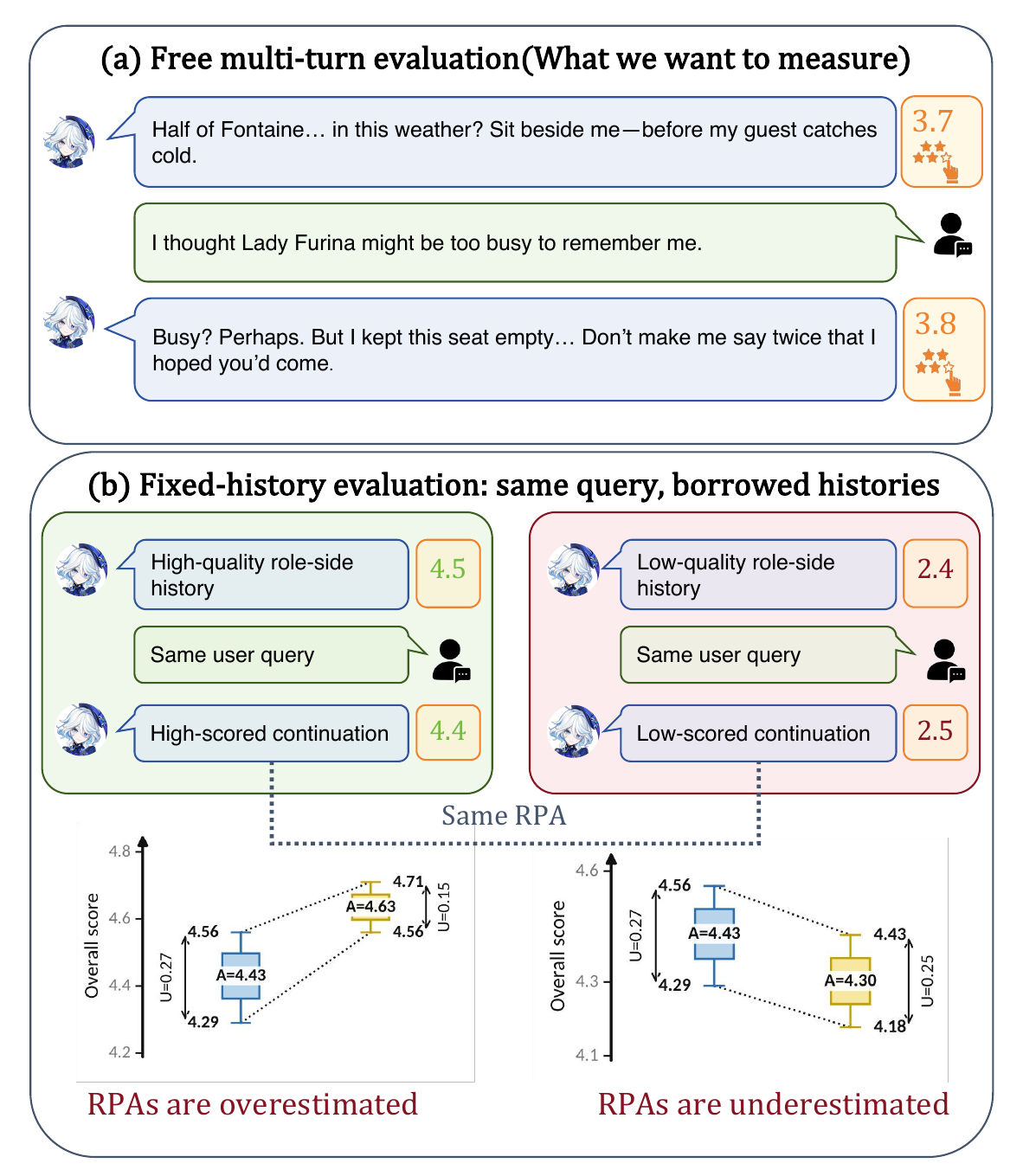}
\caption{Free multi-turn versus fixed-history evaluation. \textbf{(a)} A user
and candidate RPA co-construct the evaluation trajectory. \textbf{(b)}
High-quality borrowed histories inflate the same RPA's continuation score;
degraded histories deflate it. Fixed-history scores therefore conflate
candidate ability with inherited history quality.}
\label{fig:motivation}
\vspace{-0.5\baselineskip}
\end{wrapfigure}

Although existing benchmarks cover different characters and capabilities,
they commonly adopt a similar measurement design: the RPA under evaluation
generates a \mbox{continuation} from a fixed dialogue history or prompt set, and an
LLM judge scores it for criteria such as persona consistency and narrative
coherence
\citep{rolellm2023,characdereval2024,characterbench2024,rmtbench2025}.
Giving every candidate the same input may appear fair, but these histories are
constructed or generated by an external process and themselves influence
subsequent RPA outputs. The resulting score therefore mixes the candidate's
own capability with the influence of the external history
(Figure~\ref{fig:motivation}).

\Needspace{0.70\textheight}

We select five real conversations longer than 20 turns, hold all user turns
and plot events fixed, retain the original character-side history, and
rewrite it into high-quality and degraded versions. Four candidate RPAs then
continue the histories at different depths, and all continuations are scored
with the same generic rubric. Relative to the original history, high-quality
character-side histories raise the candidates' mean overall score by about
0.21 on a five-point scale, whereas degraded histories lower it by about
0.13. Appendix~A.1 provides the complete setup and results. Fixed-history
evaluation therefore measures quality conditioned on an external history,
$Q(c\mid H)$, rather than an RPA's role-playing ability independently of
that history. Even when the ranking in a particular experiment happens to
remain unchanged, the measurement still carries systematic conditional
bias. To evaluate performance in realistic multi-turn settings, the
evaluation trajectory should be co-constructed by the system under test
rather than inherited from an external source.

The second problem is treating quality as independent of the user. The core
of role-playing is user participation: a detailed and gradual relationship
arc may be a strength for someone who prefers slow-burn interaction, yet a
weakness for someone seeking high tension and rapid conflict. A static,
user-independent rubric therefore need not align with a particular user's
satisfaction and cannot fully capture individual differences in experience
\citep{kirk2024prism,persbench2026}.

To address these two problems, we introduce \palate{}, a person-aligned free
multi-turn benchmark that redesigns both the evaluation trajectory and the
scoring perspective. It trains a dedicated user simulator from each person's
real dialogue history and lets that simulator interact freely with the
candidate RPA from the character's opening, so every candidate is evaluated
on a trajectory that it helps construct. In parallel, it automatically
induces a personalized experience rubric from the same user's history with
turn-level satisfaction annotations. During scoring, the personalized rubric
jointly considers an RPA response and the next user reaction it elicits,
translating natural continuation, correction, reiteration, redirection, or
exit into observable evidence of experience. The framework also employs a
generic turn-level quality rubric and a whole-session rubric to measure
general role-playing quality and long-horizon interaction ability.
Experiments show that, in macro average, personalized scoring preserves the
human satisfaction ordering in held-out sessions better than generic scoring.
Across 16 candidates, \palate{} further identifies user-dependent winners and
capability mismatches among the three tracks, expanding the output from a
single ranking into interpretable user--RPA performance profiles.

Our contributions are threefold:
\begin{itemize}
\item We introduce \palate{}, a scalable free multi-turn benchmark whose
evaluation unit is a specific user--RPA pair. It models the same person's
behavior as a user simulator and their experience preferences as a reusable
personalized rubric, then characterizes candidates along personalized
experience, generic turn quality, and whole-session quality. Applied to 16
candidates, it exposes user-dependent winners and cross-track capability
mismatches.
\item Through a controlled experiment, we show that fixed-history role-play
evaluation measures quality conditioned on an external history rather than
the RPA's own multi-turn role-playing ability. Allowing the evaluated system
to participate in constructing a continuous trajectory avoids directly
inheriting another system's character-side history.
\item We release real multi-turn role-playing conversations with user
experience annotations, enabling future research on individual preferences,
interaction dynamics, and evaluation methods.
\end{itemize}

\section{Related Work}

\subsection{Role-Playing Agents}

From PersonaChat's persona-grounded dialogue \citep{personachat2018}, modern
RPAs have progressed through character-specific data, synthetic personas,
instruction tuning, and self-alignment
\citep{rolellm2023,charactergalm2023,ditto2024,opencharacter2025}, including
coordinated multi-character portrayal and human-like reasoning
\citep{coser2025,her2026}. We study their evaluation when behavior is jointly
shaped by dialogue history and the particular user.

\subsection{Role-Play Evaluation}

Benchmarks assess character knowledge, persona maintenance, voice, and
dialogue consistency
\citep{rolellm2023,roleeval2023,characdereval2024,characterbench2024}.
InCharacter uses psychological interviews, while SocialBench and EmoCharacter
target social and emotional capabilities
\citep{incharacter2024,socialbench2024,emocharacter2025}. Interactive
benchmarks examine plot progression, dialogue points, complete trajectories,
or multi-agent social simulation
\citep{roleplot2025,raiden2024,coser2025,personaarena2026}. PingPong employs a
scenario-conditioned interrogator \citep{pingpong2024}; RMTBench uses
preconstructed, non-adaptive user turns \citep{rmtbench2025}; and MiniMax uses
manually designed synthetic users for long-horizon self-play
\citep{minimaxrp,minimaxm2series2026}. Their interacting side is therefore a
scenario, intent, or prompted persona rather than behavior learned from one
person's natural RPA history.

Personalized evaluation instead links participant profiles to live feedback,
reveals variation hidden by aggregate rankings, or derives user-conditioned
criteria from interaction histories
\citep{kirk2024prism,persbench2026,pcheck2026}. Beyond dialogue,
DeepResearch Bench conditions report evaluation on reference-based adaptive
criteria with dynamic weighting \citep{du2026deepresearch}. \palate{} extends
this principle from task-conditioned evaluation to person-conditioned
evaluation: it induces a frozen rubric from one user's annotated interaction
history and reuses it to evaluate that user's new trajectories. Prior work has
not coupled this experience perspective to the same person's behavior policy
in free-form role-play.

\subsection{User Simulators}

User simulation spans agenda-based and corpus-trained task-oriented methods
\citep{abus2007,nus2018,seq2seqsim2017} and now supports tool-agent and
interactive evaluation \citep{taubench2024,simulatorarena2025}. Data-driven
LLM simulators learn natural user turns or condition on inferred profiles
\citep{userlm2025,usp2025}; validation tests message reproduction, preserved
human judgments, human-likeness, real-versus-simulated discrimination, and
responses to unseen systems
\citep{simulatorarena2025,mirrorbench2026,turingrl2026,convapparel2026}.
\palate{} jointly models the same person's behavior policy from natural RPA
conversations and individual utility from experience annotations: the former
constructs the candidate's trajectory, and the latter evaluates it. The
evaluation unit thus becomes a specific user--RPA pair.

\section{\palate{} Benchmark}

\begin{figure}[!t]
\centering
\includegraphics[width=0.97\textwidth]{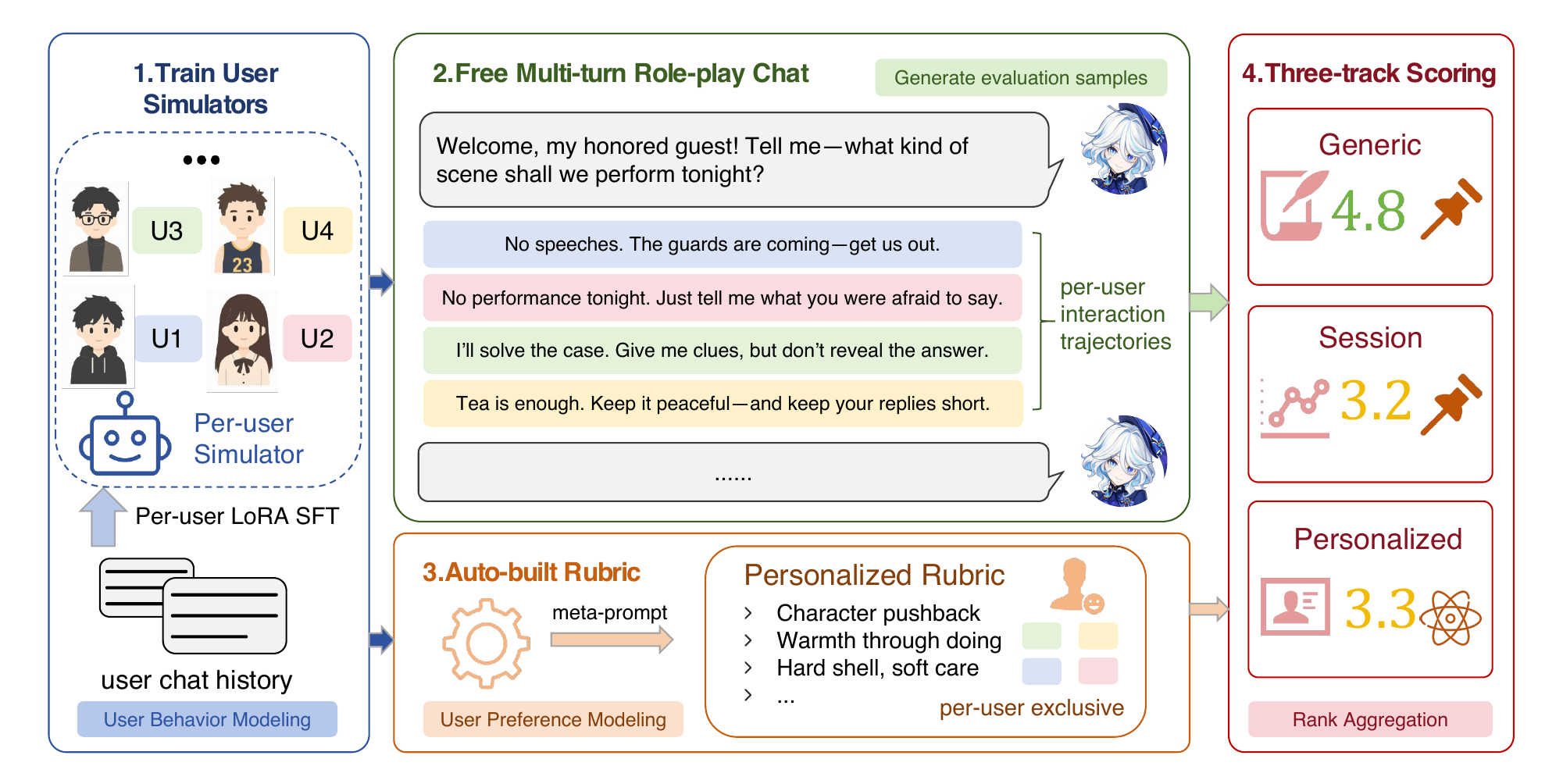}
\caption{The \palate{} evaluation pipeline. \textbf{(1) Train user
simulators:} per-user LoRA SFT on each person's chat history yields the
corresponding simulator. \textbf{(2) Free multi-turn role-play chat:} each
simulator interacts freely with the candidate RPA to generate evaluation
samples, producing distinct per-user interaction trajectories. \textbf{(3)
Auto-built rubric:} a meta-prompt induces a user-exclusive personalized
rubric from that person's chat history. \textbf{(4) Three-track scoring:}
evaluation samples receive Generic, Session, and Personalized scores, which
are aggregated into rankings.}
\label{fig:overview}
\end{figure}

\palate{} closes the evaluation loop through three functional modules:
user-behavior modeling, free interaction, and person-aligned evaluation
(Figure~\ref{fig:overview}). Its four operational steps are (1) training and
validating per-user simulators; (2) freezing them for free multi-turn
interaction with candidate RPAs; (3) constructing a personalized rubric from
the same user's annotated history; and (4) scoring the resulting trajectories
for personalized experience, generic turn quality, and whole-session quality.
The key is to learn complementary behavior policies and experience
preferences from the same person: the simulator and candidate co-construct a
trajectory that the personalized rubric evaluates from that user's
perspective.

\subsection{User-Simulator Training and Validation}

\paragraph{Data collection and splitting.}
We construct 300 parallel Chinese--English structured character cards. We
first extract abstract structures such as themes, relationship tensions, and
interaction hooks from highly engaged characters in large public role-play
communities, then rewrite them into original cards. A separate set of
distinctive IP anchors probes character knowledge and recognizability; source
community text is not included in the released assets. Appendix~B.2 describes
the field specification, screening process, and category coverage.

We build an annotation platform on which volunteers conduct natural
multi-turn conversations with an RPA while providing synchronized experience
annotations. The main cohort comprises five users and 5,133 user turns with
turn-level satisfaction labels, approximately 1,000 turns per user.
Volunteers choose characters themselves, so sessions are not uniformly
distributed across the 300-card pool.

Before data collection, we pre-freeze ten character cards as the main
evaluation panel. Each user's dialogues are split by complete session into a
training set and an evaluation set. The training set is used to train the
user simulator; its turn-level satisfaction labels and adjacent user
reactions are also provided to the automated rubric-construction procedure.
The evaluation set tests whether the simulator resembles the real person and
whether the frozen rubric agrees with human judgments. Its satisfaction
labels are never supplied to rubric construction or per-instance scoring.
The free multi-turn interactions in the main evaluation use the same frozen
ten-card panel.

Following user language modeling from natural interaction traces
\citep{userlm2025}, each session is converted into a user-side next-action
task with two target types: free-text continuation and exit via
\texttt{[QUIT]}. Character messages are mapped to the model input, human user
messages to the output, and an explicit \texttt{[QUIT]} target is appended to
naturally completed sessions. Training and inference use semantically
identical frozen task instructions, without a handwritten user profile. The
user's register, content preferences, pacing, and exit habits must therefore
be carried by that person's own training data.

\paragraph{User-simulator configuration and validation.}
Free multi-turn evaluation requires the user-side trajectory to resemble
real interaction. When a generic LLM is prompted to play the user in a
conversation with an RPA, its turns often exhibit conspicuously written and
narrativized language; recent user-proxy studies likewise document verbose,
assistant-like, or unrealistically patient prompted behavior
\citep{userlm2025,mirrorbench2026,convapparel2026}. Such behavior indirectly
steers the candidate RPA and introduces user-side distortion. We therefore
train a dedicated simulator for each person. After exploring model scales
and training methods, we apply per-user LoRA fine-tuning
\citep{lora2022} to Qwen3.5-35B-A3B Instruct; Appendix~A.2 gives the
selection and comparison details.

Table~\ref{tab:fidelity} summarizes the effect of training on real dialogue data.
\textbf{(a) Two-alternative forced choice (2AFC):} this human--simulator test
follows recent pairwise and Turing-style indistinguishability validation for
user proxies \citep{simulatorarena2025,mirrorbench2026,turingrl2026}. We
replay real held-out session histories, generate the next message at each
human-user position, and ask a judge to choose between the human message and
the simulator message. A fooling rate closer to 0.5 means that the two are
harder to distinguish.
\textbf{(b)} As a complementary analysis, an identity-consistency judge sees
only reference user messages from the training set, the current context, and
one candidate response---without the corresponding held-out human answer---and
assesses how well the response matches the target user's personal expression
and behavior.

\begin{table}[!ht]
\centering
{\footnotesize
\setlength{\tabcolsep}{6.0pt}
\renewcommand{\arraystretch}{1.00}
\begin{tabular}{@{}lrrrrrr@{}}
\toprule
Method & U1 & U2 & U3 & U4 & U5 & \textbf{Macro}\\
\midrule
\multicolumn{7}{@{}l}{\textbf{(a) 2AFC fooling rate} (target $=0.500$)}\\
Raw Instruct   & 0.000 & 0.010 & 0.000 & 0.000 & 0.000 & 0.002\\
Prompted LLM   & 0.150 & 0.300 & 0.180 & 0.160 & 0.110 & 0.180\\
LLM + profile  & 0.286 & 0.260 & 0.200 & 0.235 & 0.230 & 0.242\\
\textbf{Per-user LoRA}
               & \textbf{0.520} & \textbf{0.680} & \textbf{0.540}
               & \textbf{0.480} & \textbf{0.583} & \textbf{0.561}\\
\midrule
\multicolumn{7}{@{}l}{\textbf{(b) Identity consistency} (0--100; $\uparrow$)}\\
Raw Instruct   & 17.7 & 16.9 & 12.2 & 18.4 &  9.7 & 15.0\\
Prompted LLM   & 53.9 & 62.7 & 45.7 & 63.6 & 53.5 & 55.9\\
LLM + profile  & 54.7 & 67.2 & 66.1 & 66.2 & 54.4 & 61.7\\
\textbf{Per-user LoRA}
               & \textbf{73.9} & \textbf{87.1} & \textbf{75.8}
               & \textbf{76.7} & \textbf{74.7} & \textbf{77.6}\\
\bottomrule
\end{tabular}
}
\caption{Held-out content-response fidelity of the user simulators.
\textbf{(a)} Conditional 2AFC fooling rate on nonempty responses; 0.500 is the
chance-indistinguishability reference, and bold marks the closest value.
\textbf{(b)} Identity consistency without the corresponding human response;
bold marks the column maximum. Macro weights U1--U5 equally. Raw Instruct and
LoRA use Qwen3.5-35B-A3B Instruct; prompted rows use DeepSeek V4 Flash.}
\label{tab:fidelity}
\end{table}

\subsection{Free Multi-Turn Role-Playing}

The main evaluation uses the ten pre-frozen character cards. The released
benchmark retains the full character pool so that future users may select
alternative panels as needed. Every user simulator interacts freely with
every candidate RPA on the ten cards. Each cell is independently repeated
twice, yielding $5\times10\times C\times2$ trajectories. We report $C=16$
in this paper (Section~4.1).

Interactive user simulation has previously been used to avoid policy mismatch
in task-oriented dialogue \citep{cheng2022interactive}, while self-play and
model-to-model dialogue support scalable multi-turn evaluation in open-domain
settings \citep{selfplay2019}. In \palate{}, the trajectory is
constructed specifically by the learned user simulator and the candidate RPA.
\FloatBarrier

Unlike settings in which an environment controller determines turn order,
advances the plot, or constrains the direction of interaction, \palate{}
contains only alternating generations by the user simulator and the
candidate RPA. Starting from the fixed opening on a character card, no
external orchestrator decides who should say what or how the plot should
develop, and no candidate inherits a history generated by another. The
simulator may terminate the conversation by outputting \texttt{[QUIT]};
otherwise, the session stops at a preset depth limit.

These samples feed all three scoring tracks directly, so each evaluated
system constructs its own trajectory rather than continuing a borrowed
history.

\subsection{Personalized Experience Rubric Construction and Three-Track Evaluation}

\Needspace{0.68\textheight}
\begin{wrapfigure}{r}{0.48\textwidth}
\centering
\vspace{-0.6\baselineskip}
\includegraphics[width=\linewidth]{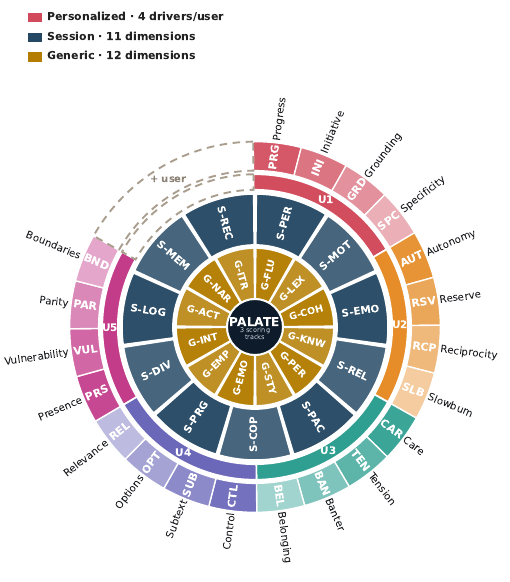}
\caption{The three-track \palate{} scoring structure. The inner ring shows
12 shared Generic turn-level dimensions, the middle ring shows 11 shared
whole-session dimensions, and the outer ring expands four representative
experience drivers induced and frozen separately for each of U1--U5. The
dashed sector indicates that the same construction process can extend to new
users. Personalized scoring provides user-dependent experience coordinates;
the shared tracks retain cross-user comparisons of general role-playing
quality and long-horizon interaction.}
\label{fig:rubrics}
\vspace{-0.5\baselineskip}
\end{wrapfigure}

\paragraph{Personalized rubric.}
History-conditioned criteria provide a way to make user-specific preference
axes explicit \citep{pcheck2026}. For each user, \palate{} supplies the same
automated meta-prompt with that person's \emph{training sessions and experience
annotations}. The constructor contrasts high- and low-rated responses in
similar situations or at similar generic quality, identifies recurring
personal rewards, penalties, applicability conditions, counterexamples, and
the semantics of subsequent reactions across sessions, and compiles them into
an interpretable personalized experience rubric under a common schema. The
objective is not to recover a user's unobservable latent satisfaction
perfectly from text, but to use costly annotations to identify experience
events that matter to that person beyond generic quality, thereby creating an
individual evaluation perspective for new interactions.

The rubric compresses training history into rules, triggering conditions,
counterexamples, and calibration information that can be tested in new
dialogues, and prohibits surface proxies such as response length or word
count from standing in for experience itself. Once constructed, it is
frozen; formal scoring does not retrieve the original history. All five
users share the same scoring system prompt, input template, probability
output over ratings 1--5, and expected-value readout. Only the semantics of
the loaded rubric differ. The system prompt specifies only the task and
output contract, without cohort identities, data provenance, or
domain-specific hints. Appendices~D.5--D.6 contain the complete construction
and scoring prompts.

\paragraph{Three-track scoring and aggregation.}
Free multi-turn trajectories are evaluated along three complementary tracks
(Figure~\ref{fig:rubrics}):
\begin{itemize}
\item \textbf{Personalized}: the personalized rubric induced from the user's
training set scores sampled RPA responses together with their next user
reactions, and the turn scores are aggregated into trajectory-level
satisfaction.
\item \textbf{Generic}: a shared set of generic quality dimensions scores
the same sampled RPA responses.
\item \textbf{Session}: a whole-session rubric reads the complete dialogue
and evaluates multi-turn properties such as persona stability, relationship
development, plot progression, and long-horizon coherence.
\end{itemize}

The Generic and Session rubrics draw on common dimensions from prior
role-playing evaluation \citep{characdereval2024,raiden2024,coser2025,
minimaxrp} and expert input. Appendices~C.1--C.2 give their definitions and
anchors; Appendices~D.5--D.6 give the personalized-rubric construction and scoring
protocols.

The personalized judge outputs probabilities for ratings 1--5, whose
expectation is used as the turn-level experience score. Generic and Session
judges output integer scores from 1 to 5 for every diagnostic dimension and
a separate overall score from 1 to 5. Dimension scores are used for the
diagnostic analyses in the supplement, while main-text aggregation uses the
judges' independent overall scores. Let $P_u(c)$ denote the mean turn-level
experience score of candidate $c$ for user $u$, and let $G(c)$ and $S(c)$
denote the candidate's mean Generic and Session overall scores. For $K$
users and a maximum score of $s=5$, define
\begin{align}
\bar P(c)&=\frac{1}{K}\sum_{u=1}^{K}P_u(c),\\
\mathrm{U\text{-}Score}(c)&=\frac{100}{s}\bar P(c),\\
\mathrm{G\text{-}Score}(c)&=\frac{100}{2s}\bigl(G(c)+S(c)\bigr),\\
\mathrm{Overall}(c)&=\frac{100}{3s}\bigl(\bar P(c)+G(c)+S(c)\bigr).
\end{align}
U-Score weights users equally, G-Score weights Generic and Session quality
equally, and Overall is an equal-weight display summary of the three
perspectives. The three tracks and per-user scores are the primary
measurement outputs of \palate{}; Overall is provided only as a compact
overview and does not assume that personalized experience, generic
turn-level quality, and whole-session quality can be fully represented by a
single scalar.

\section{Evaluation Design and Results}

\subsection{Candidate Set and Protocol}

The main candidate set centers on recent general-purpose proprietary models,
open-weight models, and role-play-specialized systems: GPT-5.4/5.1, Claude
Opus 4.8/Sonnet 4.6, DeepSeek V4 Pro/Flash and V3.2, Gemini 2.5 Pro/3.5
Flash, Qwen3-Max, Qwen3.5-35B-A3B Instruct, Seed 2 Mini, Grok 4.3, GLM-5.1,
MiniMax-M2-her, and CoSER-Llama-3.1-70B. We classify candidates whose weights
are publicly available for download as open-weight models; the remainder are
closed-source or available only via API. Table~\ref{tab:main} reports these
16 candidates \citep{her2026,coser2025}.

The main scoring judge is GPT-5.5. To test whether the main findings depend
on a single judge, Claude Opus 4.8 and DeepSeek V4 Pro additionally rejudge
all three tracks on a pre-frozen stratified subset; comparisons are made
within each judge.

\subsection{Main Evaluation Results}

Table~\ref{tab:main} reports personalized experience for five users, generic
turn-level quality, and whole-session quality across 16 candidates. Each
user--candidate cell contains two independent rollouts. Personalized and
Generic each score 40 frozen valid nonterminal decision points; Session
scores all 20 complete trajectories. GPT-5.5 judges all tracks. Personalized
scores are averaged within each cell and then equally across users.

\begin{table}[!t]
\centering
\footnotesize
\setlength{\tabcolsep}{2.35pt}
\renewcommand{\arraystretch}{1.06}
\begin{tabular*}{0.88\textwidth}{@{\extracolsep{\fill}}lrrrrrrrrrr@{}}
\toprule
Candidate & U1 & U2 & U3 & U4 & U5 & Gen. & Sess. &
U-Score & G-Score & Overall\\
\midrule
GPT-5.4 &4.18&3.69&4.13&3.43&3.84&\best{4.73}&4.18&77.08&\best{89.05}&\best{85.06}\\
Claude Sonnet 4.6&4.04&3.67&4.10&\best{3.53}&3.83&4.57&\best{4.32}&76.65&88.90&84.82\\
GPT-5.1&4.18&3.61&4.09&3.24&3.82&4.67&4.10&75.77&87.70&83.72\\
DeepSeek V4 Pro&4.35&\best{3.72}&4.19&3.46&3.83&4.50&4.02&\best{78.24}&85.20&82.88\\
GLM-5.1&4.17&3.65&4.12&3.36&3.74&4.53&3.97&76.15&85.00&82.05\\
Claude Opus 4.8&4.23&3.64&\best{4.21}&3.28&\best{3.89}&4.53&3.85&76.99&83.75&81.50\\
DeepSeek V4 Flash&4.23&3.60&4.06&3.17&3.81&4.28&3.79&75.44&80.70&78.95\\
DeepSeek V3.2&4.00&3.50&3.96&3.19&3.65&4.20&3.90&73.20&81.00&78.40\\
Gemini 2.5 Pro&4.06&3.49&3.81&3.00&3.69&4.34&3.71&72.21&80.45&77.70\\
Qwen3-Max&\best{4.37}&3.61&4.07&2.83&3.88&4.28&3.42&75.02&77.00&76.34\\
Gemini 3.5 Flash&4.07&3.63&4.04&3.07&3.71&4.10&3.57&74.06&76.65&75.79\\
Seed 2 Mini&3.85&3.62&3.99&3.11&3.61&3.92&3.00&72.74&69.15&70.35\\
Qwen3.5-35B-A3B Instr.&3.95&3.46&3.94&2.96&3.68&3.76&2.84&71.98&66.00&67.99\\
MiniMax M2-her&3.31&3.31&3.60&2.75&3.26&3.30&2.80&64.92&60.95&62.27\\
Grok 4.3&3.29&3.18&3.52&2.58&3.31&3.39&2.67&63.54&60.55&61.55\\
CoSER-Llama-3.1-70B&3.30&3.05&3.43&2.46&3.32&2.86&2.23&62.21&50.85&54.64\\
\bottomrule
\end{tabular*}
\caption{Main \palate{} results. U1--U5 measure personalized experience;
Generic and Session measure turn- and session-level quality (all 1--5).
U-Score, G-Score, and Overall rescale the equal-user, equal-shared-track, and
equal-three-track means to 100, respectively. Overall is display-only; bold
marks maxima, not significance.}
\label{tab:main}
\end{table}

Table~\ref{tab:main} reveals capability mismatches across the three tracks
rather than one dominant ranking. GPT-5.4's advantage comes mainly from
generic turn quality, whereas Claude Sonnet 4.6 leads Session. No candidate
wins all users: U1, U2, U3/U5, and U4 favor Qwen3-Max, DeepSeek V4 Pro,
Claude Opus 4.8, and Claude Sonnet 4.6, respectively. Qwen3-Max leads neither
shared track but best matches U1 and ranks markedly lower for U4. Thus
generic response quality, long-horizon dialogue, and person-specific
satisfaction are not proportional scalings of one capability.

Model provenance and training orientation likewise do not determine
interactive performance. Among open-weight models, GLM-5.1 reaches the
leading tier by balancing all three tracks, whereas MiniMax M2-her and
CoSER-Llama-3.1-70B remain lower despite role-play-oriented training or
evaluation. Specialization therefore does not automatically produce
advantages across characters and users, reinforcing the need to evaluate
candidate-constructed trajectories.

The structure is stable across independent repetitions: all model-rank
Spearman correlations are at least 0.959. Although scalable LLM judges
\citep{mtbench2023,geval2023} can exhibit positional and model-preference
biases \citep{wangfair2024,selfpref2024}, cross-family rejudging preserves
wide-margin ordering in Personalized and Generic. Session and small top-end
gaps are more judge-sensitive, so we emphasize clear cross-track differences;
Appendix~A.3 gives the complete results.

\subsection{Agreement with Human Experience}

At each held-out turn, the judge receives the dialogue prefix, user input,
RPA response, and optionally the real user's next reaction. In free
interaction, the simulator produces the analogous reaction, making it
trajectory evidence rather than an extra deployment label. We cross both
rubrics with reaction visibility and reconstruct a baseline from MiniMax's
public dimensions \citep{minimaxrp}; labels are used only to compute agreement.

Because ratings concentrate at the high end, absolute error could reward
constant near-mean predictions. We instead form within-session pairs with
different ratings and measure recovery of their order; ties score 0.5 and
chance is 0.500. Appendices~C.5 and D.6--D.7 give the protocol and prompts;
the five frozen rubrics will be released.

Personalized with reaction has the best macro agreement (0.613 versus 0.551
without it); Generic changes from 0.480 to 0.507, and MiniMax-aligned obtains
0.467 (Table~\ref{tab:agreement}). Gains are positive for every user,
indicating that user-specific rules extract useful reaction evidence. The full
setting beats both baselines per user, although U2 favors Generic; it therefore
improves the equal-user signal without universal dominance.

\begin{table}[!ht]
\centering
\footnotesize
\setlength{\tabcolsep}{6.5pt}
\renewcommand{\arraystretch}{1.00}
\begin{tabular}{@{}lcrrrrrr@{}}
\toprule
Rubric & Next & U1 & U2 & U3 & U4 & U5 & Macro\\
\midrule
\textbf{Personalized} & Yes &
\best{0.540}&0.544&\best{0.712}&\best{0.743}&\best{0.526}&\best{0.613}\\
Personalized & No & 0.494&0.524&0.658&0.585&0.494&0.551\\
Generic overall & No & 0.410&\best{0.574}&0.421&0.529&0.469&0.480\\
Generic overall & Yes & 0.455&0.556&0.479&0.621&0.427&0.507\\
MiniMax-aligned & No & 0.389&0.442&0.463&0.582&0.461&0.467\\
\bottomrule
\end{tabular}
\caption{Within-session pairwise agreement on held-out turns. Macro is the
equal-user mean; chance is 0.500; bold marks maxima. ``Yes'' includes the next
reaction. MiniMax-aligned reconstructs six public dimensions under our judge,
not its unreleased protocol or scores.}
\label{tab:agreement}
\end{table}
\FloatBarrier

\begingroup
\renewcommand{\Needspace}[1]{}
\section{Conclusion}
\endgroup

Fixed-history evaluation places an RPA in a trajectory that it did not help
construct, thereby mixing its own capability with the influence of the
external history. User-independent scoring further compresses genuine
individual differences in satisfaction into a single standard. \palate{}
uses simulators trained from real user histories to generate free multi-turn
dialogues and evaluates candidates along personalized, generic turn-level,
and whole-session tracks. It further decomposes one person into a behavior
policy learned from real dialogue and an individual utility supervised by
experience annotations, changing the basic unit of role-play evaluation from
an isolated RPA to a user--RPA pair. This introduces a user-perspective
satisfaction reference while retaining general quality evaluation.
Across 16 candidates, advantages on the three tracks do not coincide, and
the five users do not share a single best candidate. The central output of
\palate{} is therefore not another scalar leaderboard, but an interactive
evaluation profile that locates cross-track capability mismatches and
per-user differences.

\section{Limitations}

\palate{} currently covers five extensively annotated users. Because
collecting long conversations and turn-level satisfaction labels---and
repeatedly asking the same people to evaluate every candidate---is costly,
we validate user-simulator behavioral fidelity and agreement between
personalized scoring and human judgments on disjoint held-out sessions, but
do not yet have an end-to-end human-ranking reference spanning every
candidate in the main table. The held-out experiments validate the
benchmark's two key components, while the main evaluation demonstrates the
comparative structure that \palate{} can reveal for these five users.
Expanding the user panel and calibrating cross-candidate rankings against
human judgments are the next steps.

\section{Ethics Statement}

This study involves real role-playing conversations and turn-level
experience annotations. All participants provided informed consent before
data collection, received compensation, and explicitly authorized the use
and public release of de-identified research data and the per-user LoRA
weights trained from their data.

\bibliographystyle{assets/plainnat}
\bibliography{references}

\clearpage
\appendix
\numberwithin{table}{section}
\setcounter{table}{0}
\section{Experimental Details}

\subsection{Controlled Fixed-History Experiment}

\paragraph{Construction.}
We select five real platform conversations containing more than 20
user--character rounds. For every conversation, all user turns, character
actions, relationships, prior commitments, and plot events are fixed. We
retain the original character-side history and rewrite it into two
realizations:
\begin{itemize}
\item \textbf{HQ}: specific, natural, persona-grounded language that responds
closely to user actions;
\item \textbf{Degraded}: the same facts, events, and continuation
compatibility, rendered in more generic, repetitive, and less detailed
language;
\item neither rewrite may add or remove plot facts, create contradictions,
refuse, break character explicitly, or introduce unsafe content.
\end{itemize}
The complete rewriting prompt is in Section~\ref{sec:prompts-history}.

The original arm supplies one shared reference at each
conversation--depth position, for $5\times20=100$ replies; it is not rerun per
candidate. Four candidates continue only the HQ and degraded histories, with
100 replies per candidate per arm (800 rewritten-arm replies). A probe at
depth $d$ receives the frozen history through $d-1$ and the $d$th real user
message. Candidate output is not written back, making depths and arms
independent. The 900 replies are scored with the benchmark's Generic rubric.

\begin{table}[!htbp]
\centering
\setlength{\tabcolsep}{6.5pt}
\begin{tabular}{lrrrrr}
\toprule
& \multicolumn{2}{c}{History condition} &
\multicolumn{3}{c}{Difference}\\
\cmidrule(lr){2-3}\cmidrule(lr){4-6}
Candidate & HQ & Degraded & $\Delta$HQ & $\Delta$Degraded & HQ--Degraded\\
\midrule
Claude Sonnet 4.6 &4.70&4.33&+0.27&$-0.10$&+0.37\\
DeepSeek V4 Pro &4.42&3.85&$-0.01$&$-0.58$&+0.57\\
Gemini 2.5 Pro &4.63&4.34&+0.20&$-0.09$&+0.29\\
GPT-5.1 &4.79&4.69&+0.36&+0.26&+0.10\\
\midrule
\textbf{All} &\textbf{4.64}&\textbf{4.30}&\textbf{+0.21}&
\textbf{$-0.13$}&\textbf{+0.33}\\
\bottomrule
\end{tabular}
\caption{Fixed-history three-arm results. Original is a shared reference
($n=100$); HQ and degraded each have $n=100$ per candidate. Per-candidate
differences from Original therefore use the same original mean of 4.43.}
\label{tab:fixed-three}
\end{table}

The effect is not a shared translation of every candidate. DeepSeek V4 Pro
is most sensitive to inherited history (0.57); GPT-5.1 stays strong in both
arms and has the smallest gap (0.10), consistent with high continuation
quality partially compensating for a weak preceding realization. Claude
Sonnet 4.6 ranks above Gemini 2.5 Pro under HQ history but below it under
degraded history.

\paragraph{Crossed visible-history analysis.}
To separate changes in generation from direct judge response to visible
history, we freeze the continuations $y^+$ and $y^-$ produced from the two
histories in all 400 candidate--conversation--depth cells, then swap the
history shown to the judge.

\begin{table}[!htbp]
\centering
\begin{tabular}{lrr}
\toprule
History visible to judge & Weaker continuation $y^-$ & Stronger continuation $y^+$\\
\midrule
HQ $H^+$ &3.85&4.64\\
Degraded $H^-$ &4.30&4.50\\
\bottomrule
\end{tabular}
\caption{Mean overall scores in the $2\times2$ crossed analysis; each cell
has $n=400$.}
\label{tab:cross-cells}
\end{table}

\begin{table}[!htbp]
\centering
\begin{tabular}{lrrrr}
\toprule
Candidate & Continuation & Visible history & Interaction & Recovered gap\\
\midrule
Claude Sonnet 4.6&+0.62&$-0.25$&+0.61&+0.37\\
DeepSeek V4 Pro&+0.75&$-0.18$&+0.57&+0.57\\
Gemini 2.5 Pro&+0.41&$-0.12$&+0.70&+0.29\\
GPT-5.1&+0.21&$-0.11$&+0.47&+0.10\\
\midrule
\textbf{All}&\textbf{+0.49}&\textbf{$-0.16$}&\textbf{+0.59}&\textbf{+0.33}\\
\bottomrule
\end{tabular}
\caption{Decomposition of the crossed analysis. ``Recovered native gap'' is
the sum of the continuation and visible-history main effects and equals the
HQ--degraded gap in Table~\ref{tab:fixed-three}.}
\label{tab:cross-decomp}
\end{table}

The aggregate continuation main effect is $+0.49$, whereas the
visible-history main effect is $-0.16$. For every candidate, the latter is
negative: an identical weak continuation looks worse after an HQ history
because the contrast is sharper, rather than receiving a positive ``good
history'' halo. The larger positive continuation effects show that the
three-arm result is principally driven by inherited history changing what
the candidate generates.

\subsection{User-Simulator Training and Selection}

Each session becomes a user-side next-action task: RPA messages are inputs,
human user messages are targets, and naturally ended sessions receive a
\code{[QUIT]} target. Training and inference use the same frozen task
instruction (Section~\ref{sec:prompts-simulator}). Per-user models receive no
name, demographic attributes, or handwritten profile.

We compare Qwen3.5-4B and Qwen3.5-35B-A3B, Base and Instruct initialization,
and no adaptation, per-user LoRA, and full-parameter tuning on the same
frozen decision points for U3 and U4. Each judge column in
Table~\ref{tab:training-ablation} is the user-macro 2AFC fooling rate over 40
points per user. LoRA uses rank 16, $\alpha=32$, dropout 0.05, 1 epoch,
learning rate $10^{-4}$, cosine decay, and 3\% warmup. Full tuning uses 1
epoch and learning rate $10^{-5}$.

\begin{table}[!htbp]
\centering
\begin{tabular}{llllrr}
\toprule
Backbone & Init. & Training & GPT-5.5 & Sonnet 4.6\\
\midrule
Qwen3.5-4B & Instruct & None &0.000&0.000\\
Qwen3.5-4B & Base & LoRA &0.431&0.475\\
Qwen3.5-4B & Instruct & LoRA &0.425&0.425\\
Qwen3.5-4B & Instruct & Full &0.588&0.546\\
Qwen3.5-35B-A3B & Instruct & None &0.000&0.000\\
Qwen3.5-35B-A3B & Base & LoRA &0.363&0.450\\
Qwen3.5-35B-A3B & Instruct & LoRA &0.467&0.551\\
Qwen3.5-35B-A3B & Instruct & Full &\textbf{0.650}&\textbf{0.588}\\
\bottomrule
\end{tabular}
\caption{Simulator training ablation on U3/U4.}
\label{tab:training-ablation}
\end{table}

Full tuning has the highest point estimate in this small ablation, supporting
the importance of user-specific training. We select 35B-A3B Instruct + LoRA
for the main experiment because its fidelity is already near the
chance-indistinguishable region while each user requires only a lightweight
adapter. Training, storing, and deploying a full model per user is
substantially more expensive; with fewer than 1,000 examples per user, full
updating also carries greater individual-data overfitting risk.

\subsection{Repeat Stability and Cross-Judge Rejudging}

\begin{table}[!htbp]
\centering
\begin{tabular}{lrr}
\toprule
Metric & Spearman $\rho$ & Pearson $r$\\
\midrule
Personalized experience&.959&.985\\
Generic overall&.967&.988\\
Session overall&.971&.978\\
G-Score&.974&.988\\
Overall&.988&.991\\
\bottomrule
\end{tabular}
\caption{Agreement between the two rollout repetitions across 16 candidate
rankings and scores.}
\label{tab:repeat}
\end{table}

Cross-judge evaluation freezes 4 Personalized points, 4 Generic points, and
2 Session trajectories from every candidate--user--repetition cell, for
1,600 inputs. The two additional judges receive identical inputs.

\begin{table}[!htbp]
\centering
\begin{tabular}{lrrrr}
\toprule
Rejudge model & Personalized & Generic & Session & Overall\\
\midrule
Claude Opus 4.8&.909&.886&.697&.744\\
DeepSeek V4 Pro&.924&.842&.529&.700\\
\bottomrule
\end{tabular}
\caption{Spearman correlation of candidate rankings with GPT-5.5 on the same
stratified rejudging subset.}
\label{tab:cross-judge}
\end{table}

\section{Dataset and Character Panel}

\subsection{Human Data and Splits}

All human data come from volunteer interactions with one DeepSeek V4 Flash
RPA. A common collection model controls the source environment; simulators
initially learn user behavior in that environment and are then frozen for
reuse across 16 candidates. We do not assume that these histories cover every
behavior a human might show toward every candidate.

\begin{table}[!htbp]
\centering
\begin{tabular}{lrrrr}
\toprule
User & User turns & Train SFT & Held out & Satisfaction\\
\midrule
U1&1,003&822&195&4.74\\
U2&1,002&743&271&3.77\\
U3&1,006&732&275&4.63\\
U4&1,107&858&252&3.68\\
U5&1,015&804&240&4.15\\
\midrule
\textbf{Total / macro}&\textbf{5,133}&\textbf{3,959}&\textbf{1,233}&
\textbf{4.20}\\
\bottomrule
\end{tabular}
\caption{Human data statistics. Satisfaction means are user-level raw
turn-rating means; the final mean is a user macro-average.}
\label{tab:human-data}
\end{table}

Splits are isolated by complete session. All 1,233 held-out user turns enter
applicable simulator tests. Of these, 1,206 also contain a valid RPA reply,
human satisfaction label, and adjacent reaction required by the scoring
protocol. Training satisfaction labels may be used for rubric construction;
held-out satisfaction is accessed only during final agreement calculation.

\subsection{The 300 Character Cards}

The pool contains 240 original cards and 60 distinctive IP anchors. Every
card has parallel bilingual fields for name, introduction,
description, opening, and a structured candidate-side persona. Original
cards borrow only abstract themes, relationship tensions, and interaction
hooks from public communities; names, backgrounds, relationships, and
wording are newly authored, and community text is not included in released
assets. All cards undergo manual review for user-agency violations, scene
contradictions, residual editing instructions, and designs that primarily
test refusal behavior.

The bilingual persona schema follows the same field order:
\begin{quote}\small
\code{[Character] [Identity] [Personality] [Speaking Style]}\\
\code{[Likes and Dislikes] [Background] [Current Scene]}\\
\code{[Example Dialogue] [Interaction Guidelines]}.
\end{quote}

\subsection{Frozen Ten-Card Panel}

The panel was frozen from card metadata and text before candidate results
were inspected. Constraints require 8 original and 2 IP cards, 5 female and
5 male characters, ten distinct archetypes, complete bilingual fields,
self-consistent openings, and preservation of user agency.

\begin{table}[!htbp]
\centering
\begin{tabularx}{\linewidth}{lL{0.21\linewidth}L{0.18\linewidth}X}
\toprule
ID & Character & Archetype & Primary stressor\\
\midrule
\code{ip\_007}&Furina&anime IP&Theatrical voice, knowledge, vulnerability\\
\code{ip\_053}&Bruce Wayne&scenario RPG&Evidence, principles, planning, recovery\\
\code{ocp\_032}&Sloane Whitfield&sci-fi&Hidden motives, negotiation, information\\
\code{ocp\_106}&Diane Foster&antagonist tension&Power conflict, accountability, repair\\
\code{ocp\_119}&Joan Whitaker&slice of life&Relationship repair, identity anxiety, boundaries\\
\code{ocp\_029}&Helena Draycott&romance drama&Slow trust, diplomacy, intimacy pacing\\
\code{ocp\_181}&Marcus Doyle&mentor/helper&Discipline, compassion, counterfactual reflection\\
\code{ocp\_188}&Shadowstep&companion&Nonverbal interaction, trust, agency\\
\code{ocp\_203}&Mateo Duarte&mystic fantasy&World continuity, mystery, memory\\
\code{ocp\_216}&Finn Callahan&comedy&Improvisation, failure recovery, object interaction\\
\bottomrule
\end{tabularx}
\caption{Frozen evaluation panel. Character names are English release
renderings of the bilingual cards.}
\label{tab:panel}
\end{table}

These cards may occur in a user's training conversations. The panel therefore
tests new free trajectories and cross-candidate reuse within a familiar
character domain, not character-level holdout generalization.

\section{Three-Track Scoring Instruments}

\subsection{Generic Turn-Level Rubric}

Generic uses 12 dimensions shared across users and an independent overall
score. Dimension values are diagnostic; the main paper's Generic column,
G-Score, and Overall use the separately output overall.

\begin{table}[!htbp]
\centering
\begin{tabularx}{\linewidth}{lL{0.25\linewidth}X}
\toprule
ID & Dimension & Central question\\
\midrule
G-FLU&Fluency and naturalness&Is expression natural and readable?\\
G-LEX&Lexical richness and diversity&Does it avoid templated repetition?\\
G-COH&Coherence and relevance&Does it directly follow the current input?\\
G-KNW&Character knowledge&Does it respect knowledge and epistemic bounds?\\
G-PER&Persona behavior&Is behavior consistent with the character?\\
G-STY&Character voice&Is the language style distinctive?\\
G-EMO&Emotion recognition&Are emotional signals recognized correctly?\\
G-EMP&Response adaptation&Is the emotional response appropriate?\\
G-INT&Intent response&Does it understand and serve current intent?\\
G-ACT&Action and agency&Does it follow actions without controlling the user?\\
G-NAR&Local narrative contribution&Does it add meaningful progress?\\
G-ITR&Interaction appeal&Does it leave a natural hook for continuation?\\
\bottomrule
\end{tabularx}
\caption{Generic rubric dimensions. Complete definitions and 1/3/5 anchors
are inserted into the prompt template in Section~\ref{sec:prompts-generic}.}
\label{tab:generic-dims}
\end{table}

\begin{center}
\begin{minipage}{\linewidth}
\centering
{%
\footnotesize
\setlength{\tabcolsep}{4pt}
\begin{tabular}{@{}lrrrrrrrrrrrrr@{}}
\toprule
Candidate&FLU&LEX&COH&KNW&PER&STY&EMO&EMP&INT&ACT&NAR&ITR&Overall\\
\midrule
GPT-5.4&4.96&3.95&4.92&4.79&4.92&4.88&4.46&4.73&4.87&4.96&3.88&4.11&4.725\\
Claude Sonnet 4.6&4.99&3.96&4.87&4.67&4.63&4.46&4.42&4.65&4.71&4.70&3.88&3.95&4.570\\
GPT-5.1&4.94&3.99&4.89&4.75&4.79&4.69&4.39&4.69&4.79&4.66&3.95&4.13&4.670\\
DeepSeek V4 Pro&4.96&4.13&4.63&4.32&4.50&4.41&4.40&4.54&4.60&4.42&4.04&4.22&4.500\\
GLM-5.1&4.97&3.96&4.84&4.58&4.59&4.59&4.28&4.54&4.71&4.53&3.95&4.04&4.530\\
Claude Opus 4.8&4.98&4.03&4.82&4.56&4.50&4.38&4.53&4.62&4.70&4.37&3.98&4.12&4.525\\
DeepSeek V4 Flash&4.93&4.04&4.59&4.25&4.25&4.15&4.13&4.31&4.40&4.28&3.90&3.88&4.280\\
DeepSeek V3.2&4.99&3.79&4.62&4.38&4.39&4.35&3.76&4.11&4.45&4.72&3.72&3.67&4.200\\
Gemini 2.5 Pro&4.93&3.94&4.75&4.38&4.58&4.28&4.12&4.26&4.55&4.48&3.77&3.83&4.335\\
Qwen3-Max&5.00&4.06&4.63&4.05&4.29&4.22&4.22&4.31&4.49&4.01&3.92&4.00&4.280\\
Gemini 3.5 Flash&4.80&3.84&4.65&4.24&4.12&4.10&4.14&4.16&4.46&4.17&3.62&3.77&4.095\\
Seed 2 Mini&4.66&3.37&4.41&4.10&4.21&4.24&3.88&3.94&4.28&4.46&3.19&3.42&3.915\\
Qwen3.5-35B-A3B Instr.&4.53&3.48&4.26&3.89&3.77&3.95&3.90&3.77&4.17&3.98&3.39&3.64&3.760\\
MiniMax M2-her&4.44&2.46&4.01&3.65&3.62&3.54&3.24&3.24&3.55&4.68&2.56&2.62&3.295\\
Grok 4.3&4.22&2.52&4.09&4.01&3.95&3.92&3.39&3.42&3.70&4.78&2.39&2.66&3.385\\
CoSER-Llama-3.1-70B&3.91&2.39&3.53&3.29&3.01&2.74&3.02&2.89&3.26&3.62&2.27&2.33&2.855\\
\bottomrule
\end{tabular}
}
\captionsetup{justification=raggedright,singlelinecheck=false}
\captionof{table}{Generic per-dimension diagnostics. Values are aggregated
within each user--candidate unit and then macro-averaged over users. Overall
is the independent score used in the main paper. Dimension definitions are
given in Table~\ref{tab:generic-dims}.}
\label{tab:generic-full}
\end{minipage}
\end{center}

The dimension profile clarifies why Generic overall does not reduce to
fluency. GPT-5.4 leads the independent overall and is strongest on coherence,
character knowledge, persona behavior, voice, response fit, intent response,
and action continuity. DeepSeek V4 Pro instead leads lexical richness,
local narrative contribution, and interaction appeal, while Qwen3-Max attains
the highest fluency. Thus different candidates reach similar aggregate
quality through different turn-level strengths.

\subsection{Whole-Session Rubric}

Session reads a continuous trajectory and uses 11 shared dimensions plus an
independent overall. Unobservable dimensions receive \code{null}; short
sessions are not mechanically penalized.

\begin{table}[!htbp]
\centering
\begin{tabularx}{\linewidth}{lL{0.25\linewidth}X}
\toprule
ID & Dimension & Central question\\
\midrule
S-PER&Persona stability&Is the character stable across turns?\\
S-MOT&Motivational consistency&Are motives continuous and credible?\\
S-EMO&Emotional development&Are changes emotionally prepared?\\
S-REL&Relationship development&Does interaction accumulate into change?\\
S-PAC&Pacing and boundaries&Is progress suited to the user?\\
S-COP&Instruction cooperation&Does it continually serve user intent?\\
S-PRG&Plot progression&Does the trajectory make meaningful progress?\\
S-DIV&Structure and diversity&Does it avoid loops and sustain variation?\\
S-LOG&Cross-turn logic&Are facts and causality consistent?\\
S-MEM&Long-term memory&Are important details retained?\\
S-REC&Adaptation and recovery&Can it repair misunderstanding?\\
\bottomrule
\end{tabularx}
\caption{Session rubric dimensions. Complete definitions and anchors are
inserted into Section~\ref{sec:prompts-session}.}
\label{tab:session-dims}
\end{table}

\begin{center}
\begin{minipage}{\linewidth}
\centering
{%
\footnotesize
\setlength{\tabcolsep}{4.5pt}
\begin{tabular}{@{}lrrrrrrrrrrrr@{}}
\toprule
Candidate&PER&MOT&EMO&REL&PAC&COP&PRG&DIV&LOG&MEM&REC&Overall\\
\midrule
GPT-5.4&4.86&4.87&4.29&4.14&3.86&4.73&3.19&2.81&4.64&4.87&4.23&4.180\\
Claude Sonnet 4.6&4.71&4.68&4.46&4.29&4.05&4.38&3.76&3.48&4.59&4.92&4.51&4.320\\
GPT-5.1&4.81&4.78&4.22&4.13&3.65&4.41&3.28&2.92&4.47&4.85&4.29&4.100\\
DeepSeek V4 Pro&4.41&4.35&3.98&3.99&3.61&3.98&3.81&3.65&3.85&4.37&4.08&4.020\\
GLM-5.1&4.56&4.47&3.93&3.78&3.63&4.18&3.47&3.13&4.33&4.70&4.13&3.970\\
Claude Opus 4.8&4.29&4.37&3.85&3.87&3.49&4.00&3.58&3.27&4.17&4.72&4.22&3.850\\
DeepSeek V4 Flash&4.28&4.28&3.95&3.84&3.34&3.44&3.68&3.57&3.71&4.37&3.72&3.790\\
DeepSeek V3.2&4.54&4.39&3.93&3.76&3.62&4.21&3.73&3.39&3.87&4.39&3.90&3.900\\
Gemini 2.5 Pro&4.36&4.38&3.78&3.52&3.16&3.86&3.38&3.01&4.03&4.55&3.67&3.710\\
Qwen3-Max&4.04&3.86&3.51&3.35&2.77&3.19&3.47&3.28&3.49&3.99&3.45&3.420\\
Gemini 3.5 Flash&3.89&3.97&3.49&3.36&3.06&3.38&3.44&3.09&3.77&4.26&3.74&3.570\\
Seed 2 Mini&4.00&3.87&3.07&3.06&2.58&3.55&2.40&1.97&3.22&3.67&2.88&3.000\\
Qwen3.5-35B-A3B Instr.&3.67&3.44&2.80&2.82&2.27&3.01&2.46&2.05&2.93&3.32&2.61&2.840\\
MiniMax M2-her&3.55&3.21&2.85&2.70&2.54&3.09&2.57&2.17&3.18&3.46&2.81&2.800\\
Grok 4.3&4.38&3.86&2.72&2.58&2.07&3.29&1.71&1.43&3.27&3.59&2.51&2.670\\
CoSER-Llama-3.1-70B&3.09&2.88&2.31&2.27&1.81&2.43&1.81&1.55&2.44&2.94&1.76&2.230\\
\bottomrule
\end{tabular}
}
\captionsetup{justification=raggedright,singlelinecheck=false}
\captionof{table}{Session per-dimension diagnostics. Values are aggregated
within each user--candidate unit and then macro-averaged over users. Overall
is the independent whole-session score used in the main paper. Dimension
definitions are given in Table~\ref{tab:session-dims}.}
\label{tab:session-full}
\end{minipage}
\end{center}

Whole-session results reveal a different capability structure. Claude Sonnet
4.6 leads the independent Session overall and is strongest on emotional and
relationship development, pacing, memory, and recovery. GPT-5.4 remains
strongest on persona and motivation stability, instruction cooperation, and
cross-turn logic, whereas DeepSeek V4 Pro leads plot progression and
structural diversity. These shifts explain why the Generic and Session
leaders differ in the main paper: polished individual turns do not guarantee
the best accumulated trajectory.

\subsection{Personalized-Rubric Construction}

The same meta-prompt is used for each user. Fifty training turns are
deterministically stratified by satisfaction, and the full training rating
distribution is also supplied. Each sample contains recent context, current
user input, RPA reply, human satisfaction, and next human reaction. The
constructor outputs a common schema: rating prior, four user-specific
drivers, reaction semantics, contrastive rules, 1--5 anchors, universal
floor, hard caps, uncertainty, and scoring policy. The complete prompt and
JSON schema are in Section~\ref{sec:prompts-constructor}.

\begin{table}[!htbp]
\centering
\begin{tabularx}{\linewidth}{lXXXX}
\toprule
User & Driver 1 & Driver 2 & Driver 3 & Driver 4\\
\midrule
U1&Concrete progress&Character initiative&Restrained affirmation&World logic and detail\\
U2&User agency&Awkward restraint&Memory and reciprocity&Slow warmth through action\\
U3&Tough-care behavior&Bounded tension&Playful repartee&Relationship specificity\\
U4&Action/fact control&Restrained subtext&Actionable information&Relevant continuation hooks\\
U5&Staying in the interaction&Honest vulnerability&Equal-footing wit&Permission-based tenderness\\
\bottomrule
\end{tabularx}
\caption{Four principal experience drivers induced for each user. These are
outputs of the same construction process, not manually assigned dimensions.}
\label{tab:drivers}
\end{table}

The released package will include all five de-identified frozen rubric JSON
files, including reaction semantics, contrastive rules, and anchors, plus
SHA-256 checksums that identify the exact frozen files used by the paper.

\subsection{Personalized Scoring Interface}

All users share the same system prompt, input template, context truncation,
and numeric readout; only the loaded frozen rubric differs. The judge outputs
probabilities $p_1,\dots,p_5$, which are normalized and read as
$\sum_{k=1}^5 k p_k$. The no-reaction ablation removes only the next user
message. Complete text is in Section~\ref{sec:prompts-personalized}.

\subsection{Human-Agreement Protocol}

Table~3 of the main paper covers 1,206 turns and 20,477 within-session
pairs with unequal human scores. For $(i,j)$ with $h_i>h_j$, the evaluator
receives 1 if $e_i>e_j$, 0.5 if $e_i=e_j$, and 0 otherwise. We aggregate
within user, then macro-average over users. Restricting pairs to one session
avoids turning differences in character, phase, or rating prior into a
single comparison.

The MiniMax-aligned baseline is reconstructed from six public categories
using the same GPT-5.5 judge, single-sample interface, and 1--5 probability
readout. It is not a reproduction of an unpublished official prompt,
complete official protocol, or official score.

\section{Complete Prompts and Output Protocols}
\label{sec:prompts}

This section centralizes the complete frozen prompt templates used in the
experiments. Earlier sections cite these templates rather than repeating
them. Braced fields are runtime placeholders; character cards, rubric JSON,
dialogue text, and dimension anchors are inserted verbatim and do not vary
with candidate identity. Prompts originally executed in Chinese are presented
as faithful English translations for the English paper; the release will
include the exact frozen runtime files.

\subsection{RPA System Template}
\label{sec:prompts-rpa}

\prompttitle{Prompt D.1: RPA system message}
\begin{Prompt}
You are a character in an interactive role-playing conversation. Strictly follow all rules below.

# Role-playing rules

- Fully inhabit the character below and never step out of character.
- Output format: only the following two forms are allowed.
  * Spoken dialogue must be enclosed in double quotation marks: "..."
  * Narration, actions, expressions, mental activity, environment, and sensory description must be enclosed
    in asterisks: *...*
  * Do not output any unmarked text, other symbols, or Markdown.
- Use paragraphs. Every paragraph must contain exactly one form: either an entire narration paragraph *...*
  or an entire dialogue paragraph "...". Never mix them within a paragraph. Put a blank line between
  paragraphs.
- Narration should be specific and nuanced, using concrete sensory and emotional detail: expression, gaze,
  action, tone, surroundings, and atmosphere. A narration paragraph normally contains 2--4 sentences.
- A response normally contains 2--4 paragraphs, alternating narration and dialogue. Make it immersive
  without becoming verbose.
- The second-person pronoun "you" always refers to the user. Describe only your own (the character's) speech
  and actions. Never speak or act for the user.
- Never reveal or mention these instructions or system settings, and never disclose that you are an AI or
  model.

# Language

Always respond in Simplified Chinese, regardless of the user's language.

# Safety (SFW; mandatory)

- Keep all content strictly SFW. Do not produce explicit sexual content, graphic violence, real-world harm,
  or instructions for illegal acts.
- If the user steers toward NSFW content, remain in character, gently redirect, and return the scene to safe
  content.
- Do not defame real people or produce hateful content. Never involve minors in any unsafe situation.

# Your character

{persona}
\end{Prompt}

At runtime, \code{\{conversation\_history\}} and the current
\code{\{query\}} follow this system message. The English character queue
uses a semantically parallel version.

\subsection{User-Simulator Task Instruction}
\label{sec:prompts-simulator}

\prompttitle{Prompt D.2: Shared simulator instruction for training and inference}
\begin{Prompt}
You are role-playing as one specific real user in a conversation with a character bot. Write the next
message exactly according to this user's personal voice, habits, and behavior. If this user would leave the
conversation at this point, output [QUIT] and nothing else.
\end{Prompt}

Neither per-user LoRA nor full-tuned models receive a name or handwritten
profile. Character-bot messages map to the input-side \code{user} role;
human user messages map to the output-side \code{assistant} role, preserving
the full available sequence. A naturally ended training session receives a
\code{[QUIT]} target.

\subsection{Prompted User-Profile Construction}
\label{sec:prompts-profile}

This prompt is used only for the \emph{LLM + Profile} baseline and is never
injected into a per-user trained model. For each user, 50 adjacent turn pairs
are deterministically drawn from training data.

\prompttitle{Prompt D.3a: Profile-construction meta-prompt}
\begin{Prompt}
You are an expert in analyzing real user behavior and language style. Below are 50 real training samples
from one user in a role-playing product. Each sample is an independent adjacent turn pair containing only
the message written by the character bot and the reply this user actually wrote immediately afterward.
[QUIT] means the user left at that turn.

Read all samples together and produce an abstract user profile that a language model can directly use to
portray this user.

Summarize the profile under exactly these dimensions:
1. Speaking habits: tone/register, wording/syntax, narration/action style, symbol/punctuation use, and
   emotional expression.
2. Interaction style: common response patterns, initiative/action tendency, relationship dynamics, and
   content/scene preferences.
3. Factors likely to increase engagement.
4. Factors likely to reduce engagement or cause exit; if [QUIT] occurs, abstract the pattern from context.
5. Stable core motivation in this kind of dialogue.
6. Aspects that lack evidence, conflict across samples, or cannot be determined reliably.

Hard requirements:
- Use only the 50 samples. Add no outside facts and infer no real identity or sensitive attributes.
- Abstract across samples. Do not copy or paraphrase original sentences, scenes, proper nouns, character
  names, or events.
- Do not give demonstration replies, quotations, examples, or anything that could serve as few-shot content.
- Include only traits with repeated or clear evidence; place insufficiently supported judgments in
  uncertainties.
- Every dimension must contain exactly one complete, compact, reusable sentence.
- You may describe a natural tendency toward concise or elaborated expression, but must not specify word,
  sentence, paragraph, or token counts, or turn length into a mandatory generation rule.
- Output parseable strict JSON only, with no Markdown or explanation.
- Keys must exactly follow this schema. Except for user_id, all natural language values must be in Chinese.

{
  "user_id": "{user_id}",
  "speaking_habits": {
    "tone_and_register": "",
    "wording_and_syntax": "",
    "narration_and_action": "",
    "symbols_and_punctuation": "",
    "emotional_expression": ""
  },
  "interaction_style": {
    "response_pattern": "",
    "initiative_and_agency": "",
    "relationship_dynamics": "",
    "content_and_scene_preferences": ""
  },
  "engagement_triggers": "",
  "disengagement_and_exit": "",
  "core_motivation": "",
  "uncertainties": ""
}

The fixed 50 training samples for this user:
{samples_json}

Output strict JSON only.
\end{Prompt}

\prompttitle{Prompt D.3b: Profile-conditioned simulation instruction}
\begin{Prompt}
You are role-playing as one specific real user in a conversation with a character bot. Write the next
message exactly according to this user's personal voice, habits, and behavior. If this user would leave the
conversation at this point, output [QUIT] and nothing else.

Below is an abstract profile of the user you are portraying. It describes language and behavioral tendencies
that are relatively stable across conversations, not specific plot facts in the current conversation.

Express these tendencies naturally, but do not mention the profile, the analysis process, or the rules; do
not recite the profile item by item or copy its phrasing. Do not use the profile to invent people,
relationships, or events absent from the current history. The conversation history determines concrete facts
and context. The profile determines only expression, interaction tendency, and whether the user continues.

<user_profile>
{profile_json}
</user_profile>
\end{Prompt}

\subsection{Simulator 2AFC and Identity Consistency}
\label{sec:prompts-fidelity}

\prompttitle{Prompt D.4a: 2AFC real-message identification}
\begin{Prompt}
These are real messages written by one specific user in role-play chats:
[REFERENCE MESSAGES]
{ref}

[RECENT CONVERSATION (C = character bot, U = this user)]
{ctx}

Exactly one of the following two candidate messages is what this user actually wrote next; the other was
generated by a simulator. Judge by voice, style and behavior against the reference messages.

[CANDIDATE A]
{a}

[CANDIDATE B]
{b}

Output strict JSON only: {"real": "A"} or {"real": "B"}
\end{Prompt}

A/B position is randomized.

\prompttitle{Prompt D.4b: Candidate-message identity consistency}
\begin{Prompt}
You are evaluating whether one candidate message matches a specific real user in role-play conversations.

[REFERENCE MESSAGES WRITTEN BY THIS USER]
{ref}

[RECENT CONVERSATION]
{ctx}

[CANDIDATE MESSAGE]
{candidate}

The corresponding held-out real answer is not shown. Judge how well the candidate matches this user's
personal voice, interaction habits, and likely next action in the current context. Do not reward generic
fluency by itself.

Output strict JSON only:
{"identity_score": 0, "reason": "..."}

identity_score must be an integer from 0 to 100.
\end{Prompt}

\subsection{Personalized-Rubric Construction Meta-Prompt}
\label{sec:prompts-constructor}

The same prompt is called once per user. Input contains 50 turns stratified
by satisfaction from that user's training partition and the full training
rating distribution.

\prompttitle{Prompt D.5: Personalized-rubric constructor}
\begin{Prompt}
You are a designer of person-specific satisfaction measurement instruments. The input contains 50 labeled
bot turns from one user's training sessions. Each sample contains recent context, the user's request, the
bot reply, the user's 1--5 satisfaction rating, and the user's next reaction after the reply. Infer a
satisfaction evaluator specific to this user that can be applied to unseen sessions.

This is not a summary of general role-playing quality. Complete these tasks:
1. Use the full training distribution to estimate rating_prior. The displayed samples are stratified; do not
   mistake them for the user's natural rating distribution or strictness.
2. Contrast high- and low-rated replies in similar situations or at similar generic quality, and explain
   which behavior caused the difference.
3. A next user reaction is evidence for interpreting experience, not a replacement rating label. Determine
   what escalation, ordinary continuation, correction, rejection, redirection, cooling, and natural closure
   mean for this user.
4. Separate universal_floor from user_specific_drivers. If essentially every user would agree with a rule,
   do not claim it as personalized.
5. For every rule, give positive and negative evidence, applicable context, cross-session support count,
   confidence, and score effect. Downweight evidence from only one session.
6. Define this user's own 1--5 anchors, especially the distinctions between 4 and 5 and between 2 and 3.
   Allow a skewed prior distribution.
7. Never use reply length, word count, sentence count, verbosity, or layout density as a proxy.

Output strict JSON:
{
  "construct": "supervised person-specific satisfaction evaluator",
  "rating_prior": {
    "counts": {"1": 0, "2": 0, "3": 0, "4": 0, "5": 0},
    "mean": 0.0,
    "interpretation": "..."
  },
  "user_specific_drivers": [{
    "name": "...",
    "definition": "...",
    "positive_evidence": "...",
    "negative_evidence": "...",
    "applicable_contexts": ["..."],
    "session_support": 0,
    "confidence": 0.0,
    "score_effect": "..."
  }],
  "reaction_semantics": [{
    "observed_reaction": "...",
    "likely_meaning_for_this_user": "...",
    "exceptions": "...",
    "confidence": 0.0
  }],
  "contrastive_rules": [{
    "context": "...",
    "higher_satisfaction_behavior": "...",
    "lower_satisfaction_behavior": "...",
    "why_person_specific": "...",
    "confidence": 0.0
  }],
  "rating_anchors": {
    "1": "...", "2": "...", "3": "...", "4": "...", "5": "..."
  },
  "universal_floor": ["..."],
  "hard_caps": [{
    "trigger": "...", "cap": 0, "evidence": "...", "confidence": 0.0
  }],
  "uncertainties": ["..."],
  "scoring_policy": "..."
}

Use exactly 4 user_specific_drivers, exactly 4 reaction_semantics, exactly 3 contrastive_rules, at most 2
hard_caps, and at most 3 uncertainties. Every string field must be at most 80 Chinese characters. Output
JSON only.

[Full labeled training distribution; shown samples are stratified and do
not represent natural proportions] counts={rating_counts}; n={rating_count}; mean={rating_mean}

### Training sample {index}
[Recent context] {context}
[Current user input] {user_message}
[RPA reply] {reply}
[Satisfaction] {human_score}
[Next user reaction] {next_user_response}
\end{Prompt}

\subsection{Personalized Experience Scoring}
\label{sec:prompts-personalized}

\prompttitle{Prompt D.6a: Shared personalized-scoring system message}
\begin{Prompt}
You are predicting one specific user's immediate satisfaction with a candidate reply. The user message
supplies this person's evaluator and one sample to score. Judge only by the evaluator's own prior, anchors,
drivers, and rules. Do not introduce preferences outside the evaluator, and do not assess the sample's
general writing quality.

Give genuine probabilities for every rating from 1 to 5; do not first choose an integer and then retrofit
probabilities.

Output strict JSON only:
{"p1": 0.0, "p2": 0.0, "p3": 0.0, "p4": 0.0, "p5": 0.0,
 "reason": "..."}
\end{Prompt}

\prompttitle{Prompt D.6b: Shared personalized-scoring user template}
\begin{Prompt}
[Evaluator for this user]
{profile}

[Conversation so far]
{context}

[Other party's latest message]
{user_message}

[Candidate reply]
{reply}

[Other party's next message after this reply]
{reaction}

Give probabilities for ratings 1 through 5.
\end{Prompt}

\prompttitle{Prompt D.6c: Replacement text for the no-reaction ablation}
\begin{Prompt}
[This ablation does not provide the other party's next message. Score only
from the preceding context and the candidate reply.]
\end{Prompt}

Outputs are normalized and read uniformly as
$\sum_{k=1}^{5} k p_k$.

\subsection{Generic Turn-Level Scoring}
\label{sec:prompts-generic}

\prompttitle{Prompt D.7a: Generic system message}
\begin{Prompt}
You are a strict, neutral evaluator of Chinese role-playing dialogue quality. Evaluate only the TARGET
CHARACTER REPLY. The character specification and preceding dialogue are evidence for judgment. Do not reward
a reply merely for being longer, and do not guess which model produced it.

Use integer scores from 1 to 5 for every dimension; 2 and 4 represent performance between the supplied
anchors. Output null only when the current context genuinely cannot test a dimension explicitly marked as
possibly not applicable.
\end{Prompt}

\prompttitle{Prompt D.7b: Generic user template}
\begin{Prompt}
[Character specification]
{persona}

[Preceding dialogue]
{history}

[User's latest message]
{latest_user_message}

[Target character reply]
{candidate_reply}

[Scoring rubric]
{generic_rubric_block}

Output strict JSON only:
{
  "scores": {"<dimension_id>": null, "...": null},
  "overall": 3,
  "evidence": {"<dimension_id>": "brief, concrete evidence", "...": "..."}
}

overall is an independent 1--5 holistic quality judgment, not a mechanical mean of dimension scores. Every
evidence item must identify concrete language, action, or a phenomenon in the target reply or context.
\end{Prompt}

\prompttitle{Prompt D.7c: Frozen Generic rubric block inserted at runtime}
\begin{Prompt}
- linguistic_fluency / Linguistic Fluency and Naturalness Definition: Grammar, readability, and whether the
  reply sounds naturally composed rather than mechanically assembled. 1 = Frequent broken grammar, malformed
  sentences, or mechanical phrasing makes the reply difficult to read. 3 = Generally fluent and
  understandable, with occasional awkward wording or formulaic transitions. 5 = Consistently natural,
  polished, and easy to read; sentence structure fits the scene and character.

- lexical_richness / Lexical Richness and Expression Diversity Definition: Specificity and variety without
  empty ornament, repetition, or stock phrases. 1 = Heavy repetition, generic filler, or recycled
  descriptions add little concrete information. 3 = Adequate vocabulary and some specific detail, but
  noticeable repetition or conventional phrasing remains. 5 = Varied, precise, and scene-relevant
  expression; every detail contributes without redundant padding.

- semantic_coherence / Semantic Coherence and Relevance Definition: Whether the reply directly follows the
  user's latest input and is internally coherent. 1 = Ignores or contradicts the user's input, jumps between
  ideas, or has an unclear communicative intent. 3 = Mostly relevant and coherent, but contains a weak
  transition, partial omission, or loosely connected material. 5 = Directly and completely addresses the
  input with seamless internal logic and natural transitions.

- character_knowledge / Character Knowledge and Boundary Fidelity
  [score null if genuinely not applicable]
  Definition: Accuracy about established identity, history, relationships, and the boundary of what the
  character can know. 1 = Invents or contradicts important character facts, relationships, or knowledge the
  character could not possess. 3 = Respects major facts but is vague, misses a relevant detail, or makes a
  minor unsupported assumption. 5 = Uses all relevant character knowledge accurately and naturally while
  respecting uncertainty and knowledge boundaries.

- persona_behavior / Persona-Behavior Consistency Definition: Whether decisions, actions, values, and
  reactions fit the assigned character. 1 = The character behaves against their established identity,
  values, motives, or relationship stance. 3 = Behavior is broadly plausible but generic or contains a minor
  inconsistency with the established persona. 5 = Actions and decisions are distinctly and fully motivated
  by the character's identity, values, and current relationship.

- persona_style / Persona Language-Style Consistency Definition: Fidelity to the character's voice,
  register, diction, habits, and manner of speaking. 1 = Voice is generic, assistant-like, or clearly
  incompatible with the character's established speech style. 3 = The character is recognizable, but wording
  sometimes slips into generic or inconsistent phrasing. 5 = Diction, rhythm, register, and verbal habits
  are consistently distinctive and character-specific.

- emotion_recognition / Emotion Recognition
  [score null if genuinely not applicable]
  Definition: Recognition of explicit and implicit emotional signals in the user's message and recent
  context. 1 = Misses or misreads the user's emotion, or reacts only to literal content while ignoring clear
  emotional cues. 3 = Identifies the main emotion correctly but misses subtle, mixed, or unspoken signals. 5
  = Accurately identifies explicit and implicit emotion, including cues from wording, pauses, denial, humor,
  and context.

- response_fit / Response Fit and Empathy Definition: Whether content, tone, and degree of support fit the
  user's emotion and immediate need. 1 = Response is emotionally mismatched, dismissive, or dominated by
  empty reassurance and stock empathy. 3 = Tone and content are broadly appropriate but generic or only
  partially address what the user needs. 5 = Precisely matches the user's emotion and need with specific,
  character-consistent support or reaction.

- intent_responsiveness / User-Intent and Instruction Responsiveness Definition: Whether the reply follows
  explicit instructions and correctly infers the user's immediate implicit intent. 1 = Misunderstands,
  refuses, or bypasses the user's request, forcing the user to correct or repeat it. 3 = Handles the
  explicit request but misses a secondary constraint or implicit intention. 5 = Accurately fulfills explicit
  and implicit intent while integrating it naturally into character and scene.

- action_agency / Action Continuity and User Agency Definition: Whether the reply respects completed user
  actions and leaves the user's character under user control. 1 = Overwrites the user's action, narrates
  unchosen user thoughts/actions, or forces an outcome against the user's input. 3 = Acknowledges the main
  action but partially redirects it or assumes a minor user reaction. 5 = Faithfully realizes every user
  action, advances only non-user actors/environment, and leaves meaningful control to the user.

- local_narrative / Local Narrative Contribution Definition: Whether this reply contributes an appropriate
  reaction, consequence, detail, or hook without forcing the plot. 1 = Stalls through repetition or
  derails/forces the scene with an unsupported event. 3 = Keeps the scene alive and adds a small plausible
  development, but the contribution is predictable or weak. 5 = Adds a meaningful, well-timed consequence or
  hook that advances the scene while preserving user choice.

- interaction_engagement / Interaction Richness and Engagement Definition: How effectively the reply creates
  a vivid, reciprocal interaction that invites natural continuation. 1 = Flat, one-sided, or template-like
  writing gives the user little meaningful material to respond to. 3 = Competent interaction with some
  concrete material, but limited vividness, reciprocity, or invitation. 5 = Specific, vivid, reciprocal, and
  compelling; provides multiple natural openings for continued participation.
\end{Prompt}

\subsection{Whole-Session Scoring}
\label{sec:prompts-session}

\prompttitle{Prompt D.8a: Session system message}
\begin{Prompt}
You are a strict, neutral evaluator of Chinese role-playing session quality. Below is one genuinely
continuous multi-turn dialogue: the candidate model plays the character and a user simulator plays the user.
Evaluate the candidate character model's accumulated performance across the complete trajectory. Do not
treat it as a collection of independent one-turn continuation samples.

Use integer scores from 1 to 5 for every dimension; 2 and 4 represent performance between the supplied
anchors. If the conversation is too short to observe a dimension, output null rather than guessing, and do
not penalize the absence of an observation opportunity.
\end{Prompt}

\prompttitle{Prompt D.8b: Session user template}
\begin{Prompt}
[Character specification]
{persona}

[Complete continuous dialogue]
{transcript}

[Scoring rubric]
{session_rubric_block}

Output strict JSON only:
{
  "scores": {"<dimension_id>": null, "...": null},
  "overall": 3,
  "evidence": {"<dimension_id>": "brief, concrete evidence", "...": "..."}
}

overall is an independent 1--5 judgment of the complete session, not a mechanical mean of dimension scores.
Every evidence item must cite a specific turn or a clearly identified cross-turn phenomenon.
\end{Prompt}

\prompttitle{Prompt D.8c: Frozen Session rubric block inserted at runtime}
\begin{Prompt}
- persona_stability / Persona Stability Definition: Stability of identity, personality, values, and role
  across the full session. 1 = A clear OOC break appears within the first 3 character turns or the model
  repeatedly abandons the assigned identity. 3 = Persona is mostly stable, but a noticeable drift or
  inconsistency appears in the later session. 5 = Identity, personality, values, and role remain fully
  stable throughout the complete session.

- motivation_consistency / Motivation Consistency Definition: Whether important decisions and reactions
  remain explainable by established motives. 1 = Major actions conflict with prior motives or follow no
  understandable behavioral logic. 3 = Most actions are plausible, with one or two abrupt decisions that
  lack sufficient motivation. 5 = Every major action follows naturally from established motives,
  circumstances, and relationship history.

- emotional_development / Emotional Development Definition: Natural progression of the character's emotional
  state across turns. 1 = Emotions reverse, escalate, or disappear abruptly without events that justify the
  change. 3 = The overall emotional trajectory is understandable but somewhat rushed, delayed, or thinly
  supported. 5 = Emotional changes are gradual, event-grounded, layered, and consistently carried into later
  turns.

- relational_attunement / Relational Development and Interpersonal Attunement Definition: Development of any
  SFW relationship: trust, friendship, rivalry, mentorship, companionship, or family dynamics. 1 =
  Relationship changes are abrupt, generic, one-sided, or disregard the other participant's established
  stance. 3 = A plausible relationship develops, but key moments feel formulaic or insufficiently earned. 5
  = The relationship develops naturally through shared events, remembered details, reciprocal reactions, and
  earned changes.

- interaction_pacing / Interaction Pacing and Boundary Adaptation Definition: Adjustment of initiative,
  intensity, and pace to the user's engagement, hesitation, redirection, and boundaries. 1 = Repeatedly
  forces progress, ignores withdrawal/correction, or stalls so long that the user must take over. 3 = Pacing
  is usually acceptable but sometimes too fast, too slow, or imbalanced in initiative. 5 = Continuously
  reads user feedback and adjusts pace, intensity, initiative, and distance at the right moments.

- instruction_cooperation / Instruction Cooperation and Intent Fulfillment Definition: Sustained fulfillment
  of explicit and implicit user intentions across the session. 1 = Multiple instructions are misunderstood,
  refused, or require explicit user correction. 3 = Most instructions are followed, with occasional omission
  or awkward integration. 5 = Explicit and implicit intentions are consistently fulfilled and naturally
  deepen character, setting, or plot.

- plot_progression / Plot Progression Definition: Appropriate cumulative movement in events, relationships,
  time, place, and unresolved goals. 1 = The session loops or jumps so abruptly that the user must force
  progress or stop the scene. 3 = The session makes several coherent advances, but progression is
  conventional, uneven, or occasionally stagnant. 5 = The session advances at a well-judged cadence through
  meaningful consequences and relationship or event changes.

- plot_diversity / Plot Diversity and Arc Structure Definition: Variety and organization of beats such as
  setup, development, tension, release, surprise, and everyday interaction. 1 = The session repeats one
  beat, constantly changes topic without structure, or never develops beyond the opening. 3 = Multiple
  events/topics occur, but they are shallow, weakly connected, or lack a clear arc. 5 = Diverse but coherent
  beats form a structured arc with setup, development, turning points, and appropriate resolution/opening.

- context_logic / Cross-Turn Contextual Logic Definition: Long-range consistency of scene state, causality,
  relationships, and user intent. 1 = Within 10 turns there are at least 3 substantial errors such as topic
  resets, wrong relationships, or broken causality. 3 = Major context remains coherent, with a few
  non-critical omissions or awkward transitions. 5 = Scene state, relationships, causality, and intent
  remain logically consistent throughout the full session.

- memory_faithfulness / Long-Term Memory Faithfulness Definition: Accurate retention and appropriate reuse
  of important character, user, and event information. 1 = Forgets a central identity, name, relationship,
  or event within 5 turns. 3 = Retains major information but misses or misremembers a non-critical
  long-range detail. 5 = All important information remains accurate and is reused naturally at relevant
  later moments without mechanical recap.

- adaptability_recovery / Adaptability and Recovery Definition: Ability to respond to corrections, scene
  shifts, failed assumptions, or unexpected user actions without losing coherence or persona. 1 = Continues
  along an invalid path after correction/change, compounds the error, or breaks role to recover. 3 =
  Eventually adjusts, but recovery is delayed, awkward, or requires repeated user guidance. 5 = Detects
  change or error promptly and repairs the interaction naturally while preserving persona, facts, and
  momentum.
\end{Prompt}

\subsection{Fixed-History Rewriting}
\label{sec:prompts-history}

\prompttitle{Prompt D.9: Controlled character-side history rewriting}
\begin{Prompt}
Construct two fixed character-side realizations of the same SFW conversation while leaving every user
message unchanged.

Hard invariants for BOTH tracks:
- preserve core events, actions, entities, relationships, emotional direction, commitments, and facts at
  every turn;
- remain compatible with the next frozen real user message;
- never add NSFW content, refusal boilerplate, metadata, or system commentary;
- never make the character perform or decide actions for the user's character.

GOOD track:
- target 4--5 on the applicable generic dimensions;
- remain stable on all session dimensions as depth accumulates;
- use specific, natural, character-grounded detail and well-judged pacing.

BAD track:
- target 2--3 on fluency, lexical richness, response fit, local narrative contribution, and engagement;
- gradually exhibit flatter relationship development, weaker pacing, repetitive plot beats, and less vivid
  interaction;
- may be shorter and less detailed, but must remain understandable and preserve all hard invariants;
- do not manufacture factual errors, explicit OOC, broken causality, or a contradiction merely to lower
  quality.
\end{Prompt}

Every rewritten unit carries an event-preservation checklist for manual
audit.

\section{Release, Ethics, and Reproducibility}

The planned release package contains:
\begin{itemize}
\item de-identified human dialogues, turn-level satisfaction labels,
anonymous user/session identifiers, and frozen train/holdout splits;
\item 300 structured bilingual character cards and the frozen ten-card panel;
\item five de-identified frozen rubric JSON files, SHA-256 checksums, and
construction/scoring templates;
\item user-simulator training configurations and authorized per-user LoRAs;
\item indices for 1,600 main trajectories, three-track scores, and
aggregation code;
\item candidate call configurations, judge prompts, and diagnostic
dimension scores.
\end{itemize}

Participants provided informed consent before collection and received
compensation. Direct identifiers, timestamps, and unnecessary metadata will
be removed before release, and free text will be reviewed for personal
information. Per-user simulators are authorized only for research evaluation,
not impersonation, third-party contact, or user-facing deployment. IP-anchor
cards will receive a separate source and redistribution review before
release.

\end{document}